\documentclass{ijcaArticle}
\ijcaVolume{176}
\ijcaNumber{33}
\ijcaYear{2020}
\ijcaMonth{June}

\usepackage{graphicx}
\usepackage{blindtext}
\usepackage{float}
\usepackage{svg}
\graphicspath{ {./images/} }

\begin{document}

\title{Measuring Performance of Generative Adversarial Networks on Devanagari Script}

\author{ 
	\large Amogh G. Warkhandkar \\[-3pt]
	\normalsize Vidyalankar Institute of Technology \\[-3pt]
	\normalsize Mumbai, \\[-3pt]
	\normalsize Maharashtra, India \\[-3pt]
	\normalsize amoghw2025@gmail.com \\[-3pt]
\and
	\large Baasit Sharief \\[-3pt]
	\normalsize Birla Institute of Technology and Science, Pilani \\[-3pt]
	\normalsize Hyderabad Campus, Shameerpet, \\[-3pt]
	\normalsize Telengana, India \\[-3pt]
	\normalsize baasitsharief@gmail.com \\[-3pt]
\and
	\large Omkar B. Bhambure \\[-3pt]
	\normalsize Vidyalankar Institute of Technology \\[-3pt]
	\normalsize Mumbai, \\[-3pt]
	\normalsize Maharashtra, India \\[-3pt]
	\normalsize omieblablablu2996@gmail.com \\[-3pt]
}

\terms{Deep Learning, Neural Networks, Generative Models, Computer Vision, Digital Image Processing}


\maketitle

\begin{abstract}
The working of neural networks following the adversarial philosophy to create a generative model is a fascinating field. Multiple papers have already explored the architectural aspect and proposed systems with potentially good results however, very few papers are available which implement it on a real-world example. Traditionally, people use the famous MNIST dataset as a “Hello, World!” example for implementing Generative Adversarial Networks (GAN). Instead of going the standard route of using handwritten digits, this paper uses the Devanagari script which has a more complex structure. As there is no conventional way of judging how well the generative models perform, three additional classifiers were built to judge the output of the GAN model. The following paper is an explanation of what this implementation has achieved.
\end{abstract}

\section{Introduction}

\begin{figure}
	\includegraphics[width=\linewidth]{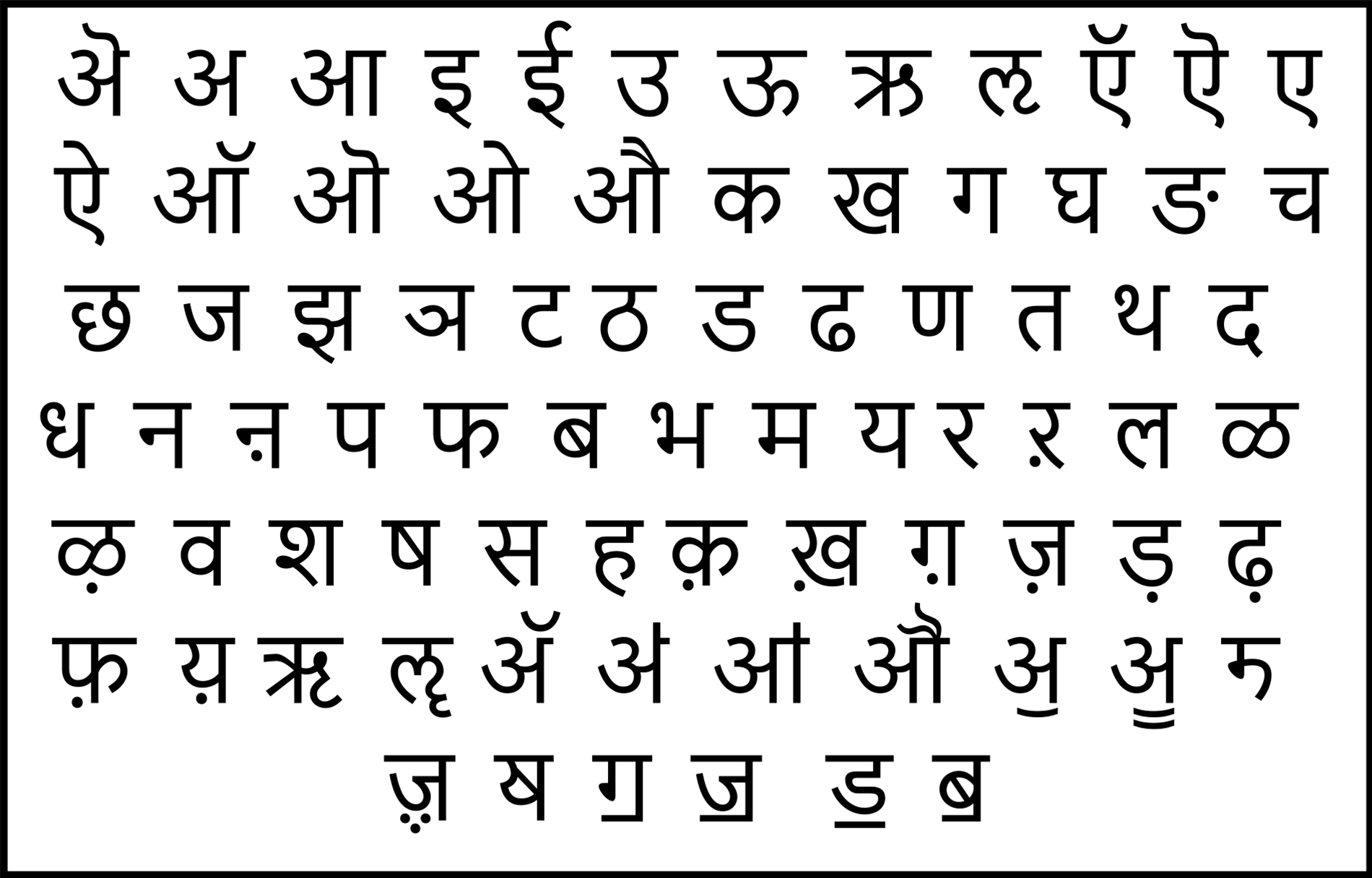}
	\caption{Devanagari script}
	\label{fig:devanagari_script}
\end{figure}

The Devanagari script, Fig. \ref{fig:devanagari_script}  is an ancient Indian script, with references in Indian culture and history back to the Vedic period. The 47 characters of the script are divided into 14 vowels and 33 consonants. Alike European languages, this script is also written from left to right. It features rounded shapes within squared outlines and a horizontal line that runs along the top of all characters.

In supervised learning, one trains the machine with well-labeled data. This allows for producing output based on previous experience. This implementation maps the input variables to an output variable and uses an algorithm to learn the relationship between them. This involves learning to predict a label associated with the data. The purpose is for the model to generalize to new data. However, in the real world, the convenience of labeled data being available is less likely.

To address this issue, there is a need for networks that can function without labeled data. GANs \cite{goodfellow2014generative} are unsupervised learning algorithms that utilize a supervised loss as part of the training. The data comes in with no labels and there is no attempt to generalize any kind of prediction to new data. The goal is for the GAN to understand what the data looks like with density estimation and generate new examples with what it has understood. 

In this implementation \footnote{All code and hyperparameters available at: \href{https://github.com/amogh-w/Paper-DevGAN}{DevGAN}.}, Devanagari script characters are fed into the GAN as input. The network will then try to generate the characters as accurately as possible. To calculate the performance metrics of the GAN, the generated characters will be tested on classifiers trained on the original dataset.

\section{Implementation}

\subsection{Generator}

\begin{figure}
	\centering
	\def\svgwidth{\linewidth}
	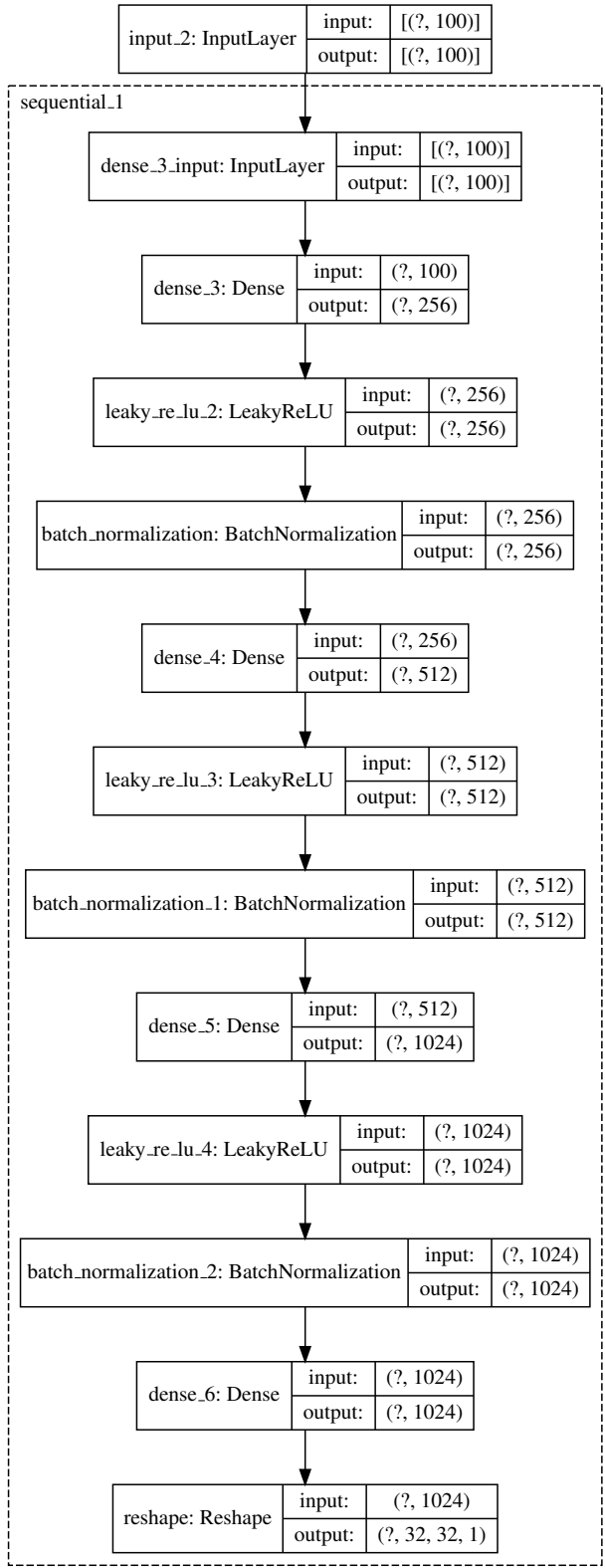
	\caption{Architecture of the Generator}
	\label{fig:generator_model}
\end{figure}

The generator model, Fig. \ref{fig:generator_model} creates new images of characters by taking a point from the latent space as input and produces a square grayscale image.

The latent space is an arbitrarily defined vector space of Gaussian-distributed values. In this implementation, the value of latent dim is set to 100. Random points from this space will be drawn and provided to the generator during training. At the end of the training, it will represent a compressed representation of a character. 

The architecture consists of three \textit{Dense} layers with the \textit{LeakyReLU} \cite{DBLP:journals/corr/XuWCL15} activation function with \textit{alpha} as 0.2 and \textit{BatchNormalization} with \textit{momentum} as 0.8 applied to every \textit{Dense} layer. The final \textit{Dense} layer is equipped with a \textit{Tanh} activation function.

\subsection{Discriminator}

\begin{figure}
	\centering
	\def\svgwidth{\linewidth}
	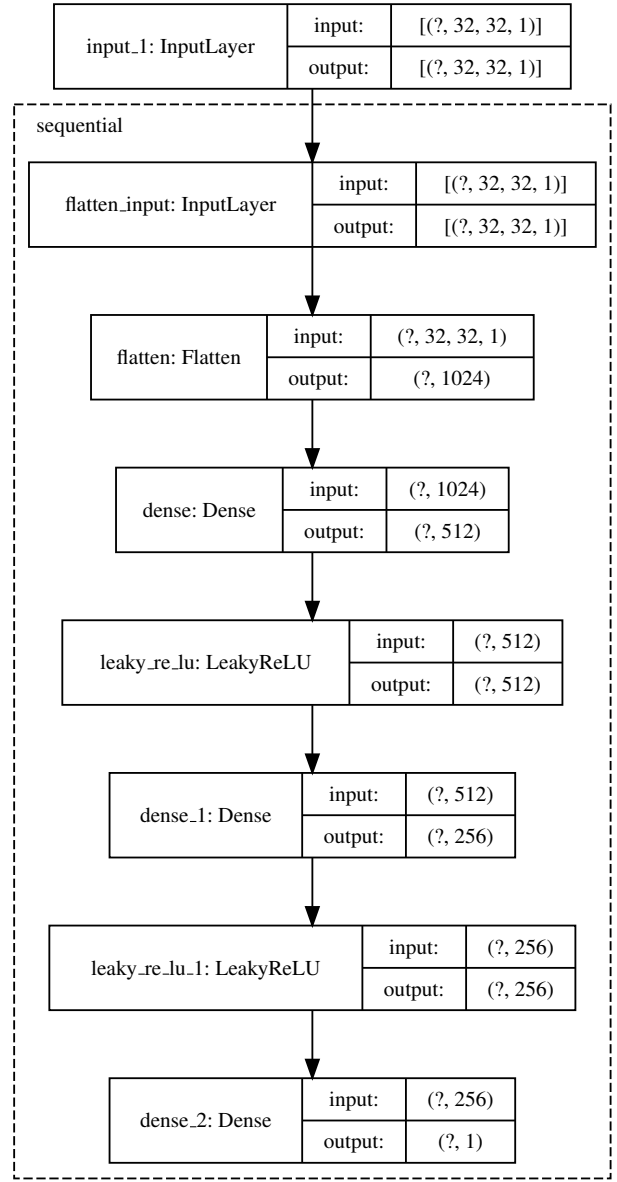
	\caption{Architecture of the Discriminator}
	\label{fig:discriminator_model}
\end{figure}

The weights in the generator model are updated based on the performance of the discriminator model. Depending upon the loss output of the discriminator, the rate at which the generator is updated is calculated. Here the adversarial relationship between these two models is defined.
The discriminator only concerns itself with the function of distinguishing between real and fake examples. Thus the layers of discriminator are marked as not trainable when combined with the generator model.

The architecture, Fig. \ref{fig:discriminator_model} consists of a \textit{Flatten} layer and three \textit{Dense} layers with \textit{LeakyReLU} activation function with \textit{alpha} as 0.2 applied to every \textit{Dense} layer. The final \textit{Dense} layer is equipped with \textit{Sigmoid} \cite{DBLP:journals/corr/abs-1811-03378} activation function.

\subsection{Classifiers}

\textbf{Classifier 1:}

It is a three-layered neural network with two layers consisting of 128 hidden units each. The layers have \textit{ReLU} \cite{DBLP:journals/corr/abs-1803-08375} activation function and the output layer has 10 units with \textit{Softmax} activation function. It uses an \textit{Adam} optimizer with loss function as \textit{sparse categorical crossentropy}.

\vspace{\baselineskip}

\textbf{Classifier 2:}

The architecture consists of three \textit{Convolutions} \cite{koushik2016understanding}\cite{liu2016better} and three \textit{Dense} layers. All the Convolutions include \textit{BatchNormalization}, \textit{ReLU} activation function, \textit{MaxPooling}, and \textit{Dropout} \cite{JMLR:v15:srivastava14a}.

The first \textit{Convolution} consists of 64 \textit{filters} of size (5,5) with \textit{same} padding along with \textit{MaxPooling} \textit{size} of (2,2) and \textit{strides} of (2,2). The \textit{Dropout} rate is 0.25. The second \textit{Convolution} consists of 32 \textit{filters} of size (3,3) with \textit{valid} padding and \textit{ReLU} activation function and continued with the same \textit{MaxPooling} and same \textit{Dropout} as above. The third \textit{Convolution} consists of 16 \textit{filters} of size (3,3) with \textit{same} padding and \textit{ReLU} activation function and continued with the same \textit{MaxPooling} and same \textit{Dropout}. The Convolutions are continued by three \textit{Dense} layers with the first two of them having \textit{Dropouts} and third being the output layer. The first \textit{Dense} layer consists of 128 units with \textit{ReLU} activation function and \textit{Dropout} with a rate of 0.25. The second \textit{Dense} layer consists of 32 units with \textit{ReLU} activation function and \textit{Dropout} with a rate of 0.5.

The last \textit{Dense} layer has 10 units with \textit{Softmax} activation as the output.  The model is compiled with \textit{Adam} \cite{kingma2014adam} optimizer with loss function as {categorical crossentropy}.

\vspace{\baselineskip}

\textbf{Classifier 3:}

This architecture is similar to Classifier 2, with hyperparameters like the number of filters and number of neurons being different and \textit{BatchNormalization} applied after every Dense layer except the output. The model is compiled with \textit{RMSprop} \cite{ruder2016overview} optimizer with the same loss function.

\section{Observations}

\subsection{Dataset}

The dataset \cite{7400041} is an image database \footnote{Dataset available at: \href{https://archive.ics.uci.edu/ml/datasets/Devanagari+Handwritten+Character+Dataset}{
Devanagari Handwritten Character Dataset}.} of Handwritten Devanagari characters. There are 46 classes of characters with 2000 samples each. The images are in png format with 32x32 resolution. The actual character is centered within 28x28 pixel and a padding of 2 pixels is added on four sides of the character.

\begin{table}[H]
\tbl{Classifier Metrics on Original Dataset}{
\begin{tabular}{|l|l|l|l|l|}
\hline
& loss   & accuracy & val\_loss & val\_accuracy \\ \hline
Classifier 1 & 0.0268 & 0.9920   & 0.5579    & 0.8856        \\ \hline
Classifier 2 & 0.2135 & 0.9415   & 0.0388    & 0.9887        \\ \hline
Classifier 3 & 0.0106 & 0.9964   & 0.1006    & 0.9808        \\ \hline
\end{tabular}}
\label{tab:original_dataset_metrics}
\end{table}

\subsection{GAN Output}

\begin{figure}[H]
\includegraphics[width=\linewidth]{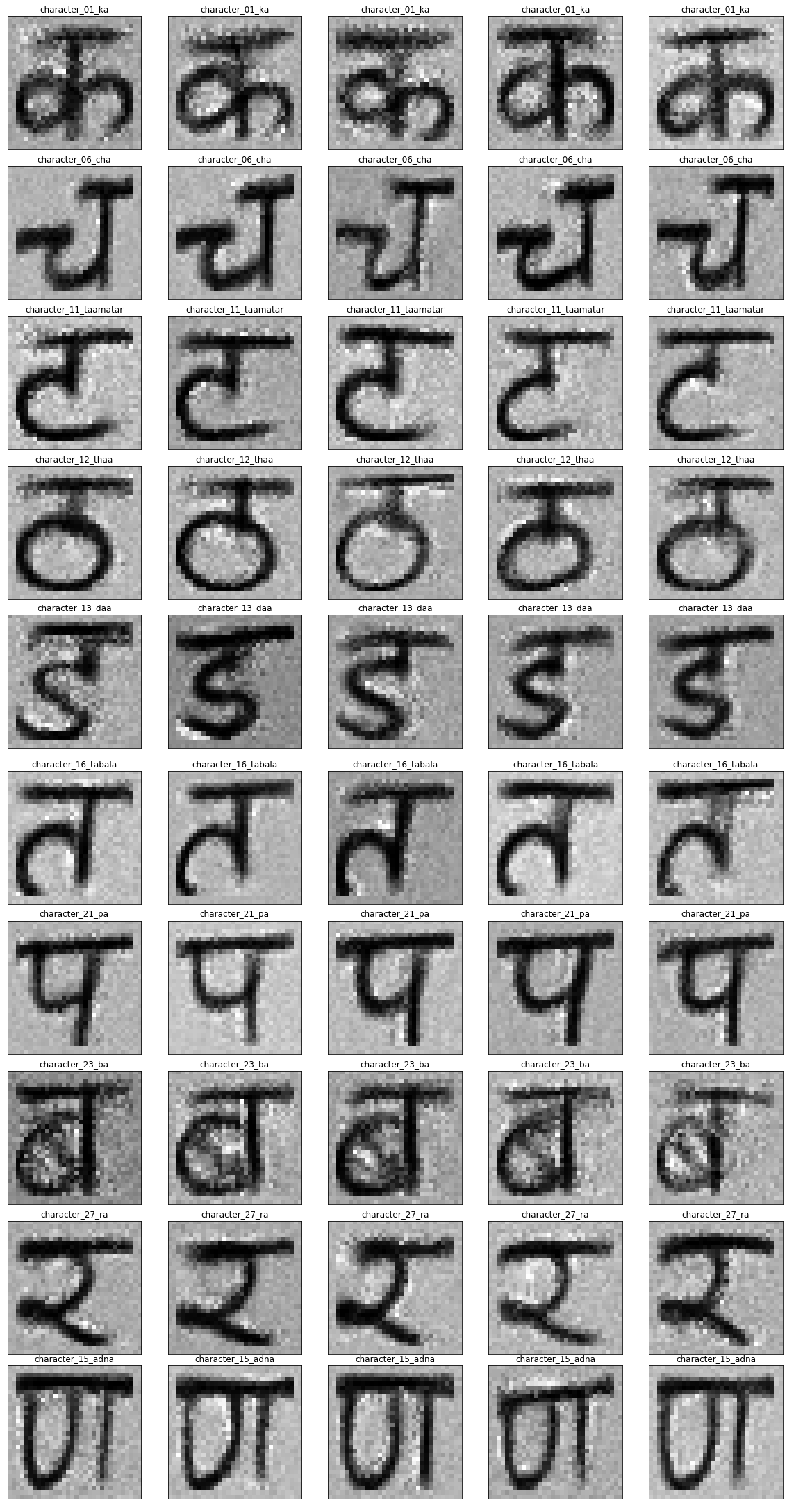}
\caption{Generated Characters}
\label{fig:gan_output}
\end{figure}

The GAN model was trained for 10,000 epochs. At every 500 epoch, the output of the model was stored. Every model started showing promising results at 5000 epochs. Training for 5000 more iterations resulted in better character generation with distinct boundaries between the character and the background. The training time was around 2 hours on an \textit{NVIDIA 1050Ti} GPU.
Despite the generated characters being readable to the human eye, Fig \ref{fig:gan_output}, a significant amount of noise was observed.

\subsection{Classifier Output}

All three classifiers performed well on the original dataset, Table \ref{tab:original_dataset_metrics}. However, on the generated characters their performance was very unsatisfactory, Table \ref{tab:generated_character_metrics}. This failure was accounted to the previously encountered noise in the generated data. Although the generated characters seemed readable to the human eyes, the classifiers were not able to differentiate the characters correctly.

\begin{table}[H]
\tbl{Classifier Metrics on Generated Characters}{
\begin{tabular}{|l|l|l|l|l|}
\hline
& loss    & accuracy \\ \hline
Classifier 1 & 11.6530 & 0.0900   \\ \hline
Classifier 2 & 21.9342 & 0.1230   \\ \hline
Classifier 3 & 8.0340  & 0.1590   \\ \hline
\end{tabular}}
\label{tab:generated_character_metrics}
\end{table}

\subsection{Generated Data Cleaning}

\begin{figure}
\includegraphics[width=\linewidth]{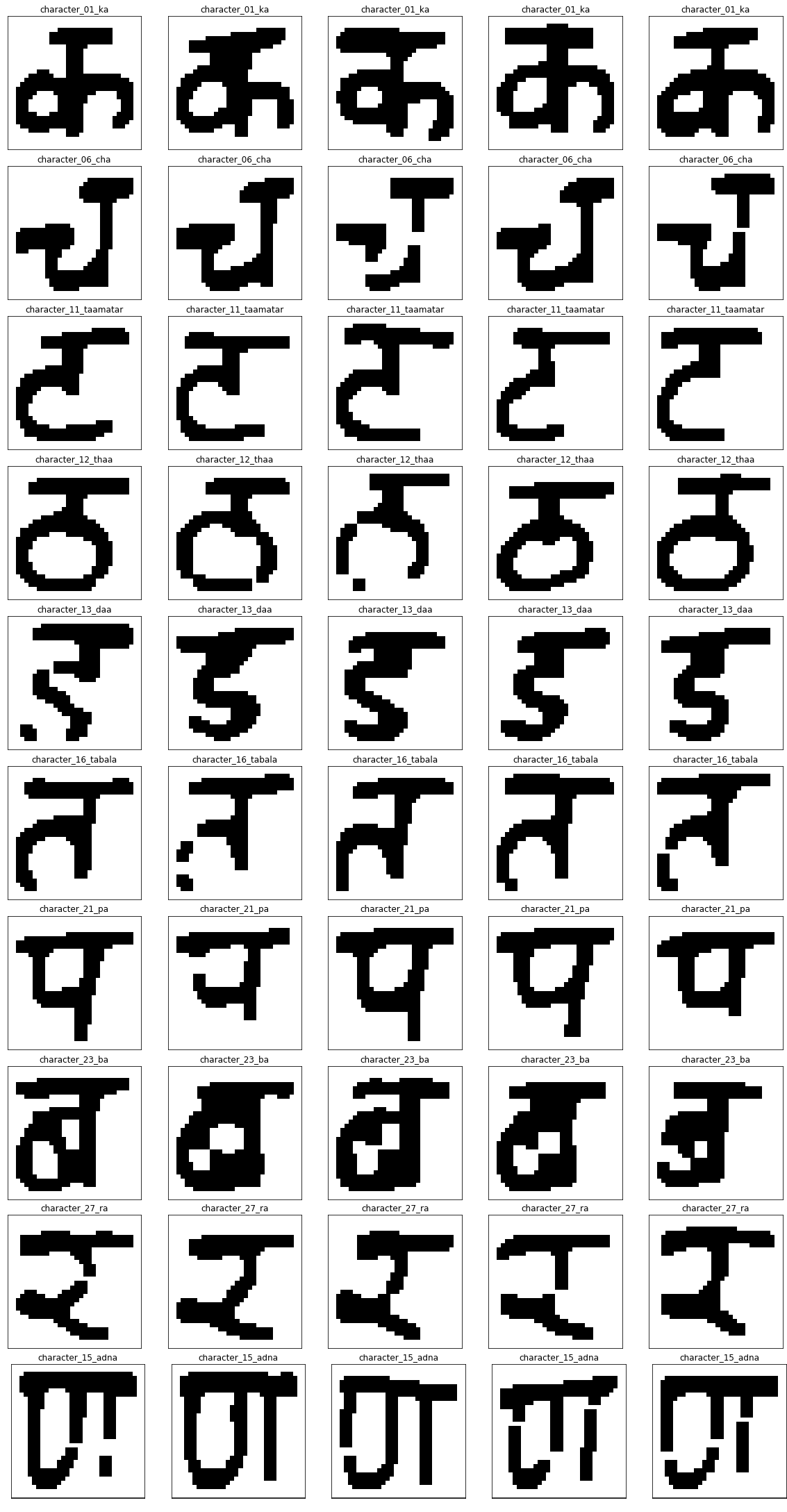}
\caption{Cleaned Generated Characters}
\label{fig:cleaned_gan_output}
\end{figure}

All the images were passed through a \textit{Gaussian Blur} \cite{gaussian} \textit{filter} of size 3x3. Furthermore, \textit{Otsu's Thresholding} \cite{otsu} was applied to segment the images into \textit{two-pixel} values i.e. 0 and 255. After this, two morphological operations \cite{morphologicalres}\cite{morphologicalop} were done i.e. opening and closing with \textit{ones} kernel of size 3x3. Finally, they were all passed through a \textit{bitwise NOT} so that the characters resembled the original data, Fig \ref{fig:cleaned_gan_output}.

\subsection{Final Classifier Output}

After the data cleaning, the result of the classifiers improved greatly. Table \ref{tab:cleaned_generated_character_metrics} contains the statistics and comparisons of the classifier outputs.

\begin{table}[H]
\tbl{Classifier Metrics on Cleaned Generated Characters}{
\begin{tabular}{|l|l|l|l|l|}
\hline
& loss   & accuracy \\ \hline
Classifier 1 & 1.0490 & 0.8930   \\ \hline
Classifier 2 & 1.7370 & 0.8692   \\ \hline
Classifier 3 & 1.2778 & 0.8660   \\ \hline
\end{tabular}}
\label{tab:cleaned_generated_character_metrics}
\end{table}

\section{Conclusions and Future Work}
Lack of data has always been a big problem in solving real-world challenges involving machine learning and deep learning applications. GANs will be beneficial to a lot of data practitioners because of its possibility of generating real-like artificial data. 

This paper has demonstrated the viability of Generative Adversarial Networks on real-life datasets. Even though the generated characters were very noisy, it is safe to assume that in the future there will be a GAN architecture that will take this anomaly into account and create significantly better output. Further experiments can be done on multiple architectures and a hybrid architecture can be developed which might result in the cleaner generation of images. Furthermore, there are high chances of more agile architectures being developed which will make the training and testing faster and more efficient.

\bibliographystyle{ijcaArticle}
\bibliography{References}

\begin{thebibliography}{10}

\bibitem{7400041}
S.~{Acharya}, A.~K. {Pant}, and P.~K. {Gyawali}.
Deep learning based large scale handwritten devanagari character recognition.
In {\em 2015 9th International Conference on Software, Knowledge, Information
  Management and Applications (SKIMA)}, pages 1--6, 2015.

\bibitem{DBLP:journals/corr/abs-1803-08375}
Abien~Fred Agarap.
Deep learning using rectified linear units (relu).
{\em CoRR}, abs/1803.08375, 2018.

\bibitem{gaussian}
Estevao Gedraite and M.~Hadad.
Investigation on the effect of a gaussian blur in image filtering and
  segmentation.
pages 393--396, 01 2011.

\bibitem{goodfellow2014generative}
Ian~J. Goodfellow, Jean Pouget-Abadie, Mehdi Mirza, Bing Xu, David
  Warde-Farley, Sherjil Ozair, Aaron Courville, and Yoshua Bengio.
Generative adversarial networks, 2014.

\bibitem{kingma2014adam}
Diederik~P. Kingma and Jimmy Ba.
Adam: A method for stochastic optimization, 2014.

\bibitem{koushik2016understanding}
Jayanth Koushik.
Understanding convolutional neural networks, 2016.

\bibitem{liu2016better}
Mengchen Liu, Jiaxin Shi, Zhen Li, Chongxuan Li, Jun Zhu, and Shixia Liu.
Towards better analysis of deep convolutional neural networks, 2016.

\bibitem{DBLP:journals/corr/abs-1811-03378}
Chigozie Nwankpa, Winifred Ijomah, Anthony Gachagan, and Stephen Marshall.
Activation functions: Comparison of trends in practice and research for deep
  learning.
{\em CoRR}, abs/1811.03378, 2018.

\bibitem{morphologicalres}
A.M Raid, Wael Khedr, Mohamed El-dosuky, and Mona Aoud.
Image restoration based on morphological operations.
{\em International Journal of Computer Science, Engineering and Information
  Technology}, 4:9--21, 07 2014.

\bibitem{ruder2016overview}
Sebastian Ruder.
An overview of gradient descent optimization algorithms, 2016.

\bibitem{morphologicalop}
Ravi Srisha and Am~Khan.
Morphological operations for image processing : Understanding and its
  applications.
12 2013.

\bibitem{JMLR:v15:srivastava14a}
Nitish Srivastava, Geoffrey Hinton, Alex Krizhevsky, Ilya Sutskever, and Ruslan
  Salakhutdinov.
Dropout: A simple way to prevent neural networks from overfitting.
{\em Journal of Machine Learning Research}, 15(56):1929--1958, 2014.

\bibitem{DBLP:journals/corr/XuWCL15}
Bing Xu, Naiyan Wang, Tianqi Chen, and Mu~Li.
Empirical evaluation of rectified activations in convolutional network.
{\em CoRR}, abs/1505.00853, 2015.

\bibitem{otsu}
Jun Zhang and Jinglu Hu.
Image segmentation based on 2d otsu method with histogram analysis.
pages 105--108, 01 2008.

\end{thebibliography}

\end{document}